\documentclass{article}

\usepackage{functions}
\usepackage{graphicx} 
\usepackage{amsmath,amsthm,amssymb,bbm,stmaryrd,bm}
\usepackage{url}
\usepackage{dsfont}
\usepackage{color}
\usepackage{wrapfig}
\usepackage{listing}
\usepackage{amsopn}

    \newcommand{\Xc}{\mathcal{X}}

\newcommand{\R}{\mathbb{R}}

\newcommand{\reals}{\mathbb{R}}

\usepackage{amsthm}

\renewcommand{\expect}[2]{\mathop{\mathbb{E}}\limits_{#1}\big[#2\big]\!\!}

\newcommand{\by}{y}

\newcommand{\normp}[1]{\norm{#1}_p}
\newcommand{\network}{g}

\newcommand{\loss}{\ell}

\newcommand{\adv}[1]{\tilde{#1}}
\newcommand{\perturb}[1]{\delta_{#1}}
\newcommand{\adveps}{\epsilon}

\usepackage{spconf,amsmath,graphicx,booktabs,lipsum,caption,threeparttable}

\def\x{{\mathbf x}}
\def\u{{\mathbf u}}
\def\v{{\mathbf v}}

\def\Xc{{\cal X}}
\def\X{{\mathbf X}}
\def\U{{\mathbf U}}

\title{Fooling end-to-end speaker verification with adversarial examples}
\name{Felix Kreuk$^1$, Yossi Adi$^1$, Moustapha Cisse$^2$, Joseph Keshet$^1$}
\address{$^1$Bar-Ilan University, Israel~~~~~~~~$^2$Facebook AI Research}
\begin{document}
%
\maketitle
\begin{abstract}
Automatic speaker verification systems are increasingly used as the primary means to authenticate costumers. Recently, it has been proposed to train speaker verification systems using end-to-end deep neural models. In this paper, we show that such systems are vulnerable to adversarial example attacks. Adversarial examples are generated by adding a peculiar noise to original speaker examples, in such a way that they are almost indistinguishable, by a human listener. Yet, the generated waveforms, which sound as speaker A can be used to fool such a system by claiming as if the waveforms were uttered by speaker B. We present white-box attacks on a deep end-to-end network that was either trained on YOHO or NTIMIT. We also present two black-box attacks. In the first one, we generate adversarial examples with a system trained on NTIMIT and perform the attack on a system that trained on YOHO. In the second one, we generate the adversarial examples with a system trained using Mel-spectrum features and perform the attack on a system trained using MFCCs. Our results show that one can significantly decrease the accuracy of a target system even when the adversarial examples are generated with different system potentially using different features.
\end{abstract}
\begin{keywords}
Automatic speaker verification, adversarial examples
\end{keywords}
%


\section{Introduction}
\label{sec:intro}

Automatic speaker verification is the task of verifying that a spoken utterance has been produced by a claimed speaker. It is one of the most mature technologies for biometric authentication deployed by banks and e-commerce as the primary means to authenticate customers online and over the phone. The vulnerability of such a system is a real threat to these applications.  

The standard verification protocol comprises the following three steps: training, enrollment, and evaluation. In the training stage,  one learns a suitable internal speaker representation from a set of utterances and builds a simple scoring function. In the enrollment stage, a speaker provides a few utterances which are used to estimate the speaker model. During the evaluation stage, the verification task is performed by scoring a new unknown utterance against the speaker model. If the resulted score is greater than a pre-defined threshold, the system predicts that the unknown utterance produced by the claimed speaker. When the authentication is based on the voice of the speaker, irrespective of what the speaker said, the system is a text-independent speaker verification system.

Most of the modern speaker verification systems have several components. For example, the combination of i-vector for speaker representation and probabilistic linear discriminant analysis (PLDA) for a scoring function has become the dominant approach, both for text-dependent and text-independent speaker verification \cite{kenny2010bayesian, dehak2011front, reynolds2000speaker}.  Recently, Heigold et al.~\cite{heigold2016end} proposed to train speaker verification systems in an end-to-end fashion using deep neural models. This approach allows to directly learn from utterances, which improves capturing long-range context and reduces the complexity (one vs. number of frames evaluations per utterance), and the direct and joint estimation, which can lead to better and more compact models. Moreover, this approach often results in considerably simplified systems requiring fewer concepts and heuristics. 

Although deep neural networks have enabled several breakthroughs in notoriously difficult problems such as image classification \cite{krizhevsky2012imagenet,he2016deep}, speech recognition \cite{amodei2016deep,graves2006connectionist}, speech processing \cite{adi2015vowel,adi2017sequence} and machine translation \cite{bahdanau2014neural}, it has been shown \cite{goodfellow2014explaining} that they are not robust to tiny perturbations in the input space. Indeed, adding a well-chosen small perturbation to the input of a network can change its prediction. When the difference between the perturbed image and the original image is indistinguishable by the human eye the example, the perturbed image is called an \emph{adversarial example}. 

Adversarial examples were first introduced in ~\cite{szegedy2013intriguing}. Their study first demonstrated that deep neural networks could achieve high accuracy on previously unseen examples while being vulnerable to small adversarial perturbations. This finding has recently aroused keen interest in the community~\cite{goodfellow2014explaining,papernot2016practical,szegedy2013intriguing,tabacof2016exploring}. Several studies have subsequently analyzed the phenomenon~\cite{shaham2015understanding,fawzi2016robustness} and various approaches have been proposed to improve the robustness of neural networks~\cite{papernot2016distillation,cisse2017parseval}. However, most of the previous works on adversarial examples are focused on the vision domain. 

In this work, we investigate the generation of adversarial examples to attack an end-to-end neural based speaker verification model. We demonstrate the generation of adversarial examples to attack a text-dependent speaker verification system while using the architecture proposed in \cite{heigold2016end}. 

To the best of our knowledge, the only work that applied adversarial attacks to speech data is \cite{cisse2017houdini} where the authors present an adversarial attack on an end-to-end automatic speech recognition system. We are unaware of any previous study on adversarial examples for fool speaker verification systems.  
A different approach for attacking speaker verification systems is known as \emph{spoofing attack} \cite{wu2017asvspoof,hadid2015biometrics}. In that type of attack, an adversary may use a falsifying voice, such as the recorded file of another person, as input for the speaker verification system. Our approach is different since our goal is to generate acoustic utterances which sound as speaker A (at least to the human ear) but can be used to fool the system by claiming that utterances were produced by speaker B.



This paper is organized as follows. In Section~\ref{sec:problem_setting} we formally set the notation and definitions used throughout the paper. Section~\ref{sec:model} provides a detailed description of the speaker verification model. In Section~\ref{sec:adv_attack} we describe the mechanism behind the generation of adversarial examples. In Section~\ref{sec:experiments} we report the results of attacking the speaker verification model in various ways. We conclude the paper with a discussion in Section~\ref{sec:conclusion}.




\section{Notations and definitions}
\label{sec:problem_setting}

In this section, we formulate the task of speaker verification rigorously and set the notation for the rest of the paper. We denote the domain of the acoustic feature vectors by $\Xc \subset \R^d$. The acoustic feature representation of a speech signal is therefore a sequence of vectors $\x = (x_1,x_2,...x_T)$, where $x_i \in \Xc$ for all $1 \leq i \leq T$. The length of the input signal varies from one signal to another. Thus $T$ is not fixed. We denote by $\Xc^*$ the set of all finite-length sequences over $\Xc$. 

Recall that in speaker verification the goal is to assert if the claimed speaker spoke the input utterance. Specifically, the speaker verification system is a function that gets as input an utterance $\x$ produced by an unknown speaker, and a set of $n$ enrollment utterances produced by speaker $k$ denoted as  $\X^k=\{\x^k_1,...,\x^k_{n}\}$. The output of the system is a real number in the simplex $p\in[0, 1]$ estimating the probability that the utterance $\x$ produced by speaker $k$.

Let $g_\theta: \Xc^* \times (\Xc^*)^n \rightarrow [0,1]$ be the speaker verification function implemented as neural network with a set of parameters $\theta$. Given a set of training examples, the parameters $\theta$ are found by minimizing the negative log likelihood loss function. Each example in the training set is a tuple $(\x,\X^k, y)$ of a spoken utterance $\x\in\Xc$, an enrollment set of a speaker $k$, $\X^k$, and a binary label $y\in\{0,1\}$ indicating whether the utterance $\x$ was produced speaker $k$. We denote by $\ell:[0,1]\times\{0,1\}\to\reals$ the log-likehood loss function.


\section{End-to-end deep network model}
\label{sec:model}

\begin{figure}[t]
\begin{minipage}[b]{1.0\linewidth}
  \centering
  \centerline{\includegraphics[width=6.6cm]{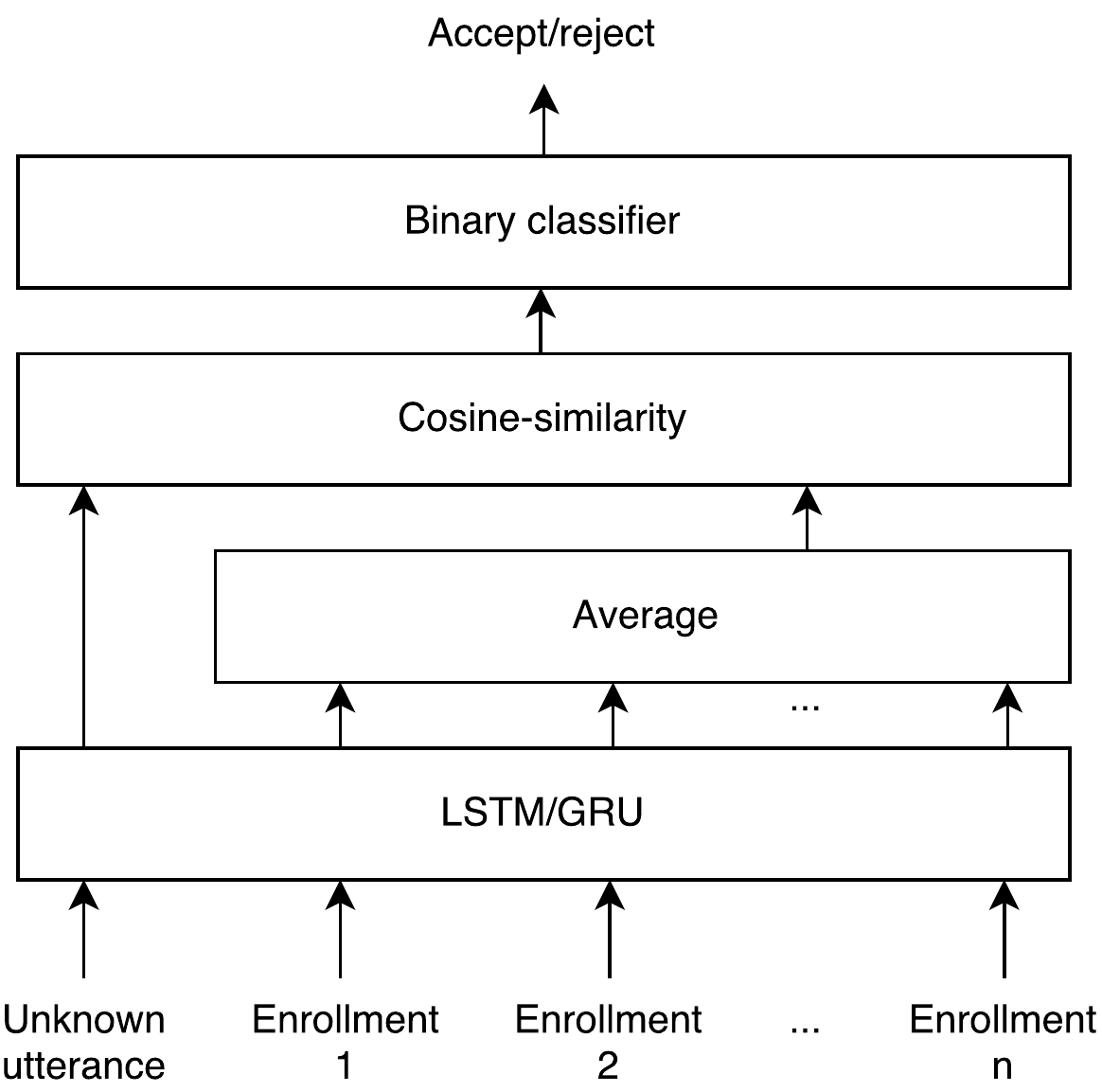}}
\end{minipage}
\caption{End-to-end deep network model for speaker verification. The architecture is based on \cite{heigold2016end}.}
\label{fig:res}
\end{figure}

In this section, we describe the network architecture used as our speaker verification function. The architecture was initially proposed in \cite{heigold2016end}, and serves as a baseline in recent work on end-to-end models for speaker verification \cite{snyder2016deep,zhang2016end}. It is depicted in Figure~\ref{fig:res}.

Recall that the input to the verification function is an unknown utterance and a set of $n$ enrollment utterances. Each of the $n+1$ utterances is fed into a recurrent LSTM network and is represented as an embedding vector of size $D$. Denote by $\u$ and by $\U^k=(\u^k_1, \ldots, \u^k_n)$ the embeddings of the unknown utterance $x$ and the enrollment set $\X^k$. The embeddings of the enrollments set are averaged to a single vector:
$$
\v^k=\frac{1}{n}\sum_{i}^{n} \u^k_i.
$$
Then the resemblance between the embedding of the unknown utterance and the average embedding of the enrollment is computed using the cosine-similarity function:
$$
sim(\u,\v^k)=\frac{\u\cdot \v^k}{\|\u\| \|\v^k\|}.
$$
The final stage takes the cosine-similarity, multiplies it by a scalar and add a bias to generate a probability estimation. One can think of this layer as an automatic setting of the threshold detection. The whole network is trained using with negative log-likelihood loss function.

\section{Generating adversarial examples}
\label{sec:adv_attack}

Given an input utterance $\x$, an adversarial example is a perturbed version of the original pattern 
$$
\adv{\x} = \x + \perturb{\x}, 
$$
 where $\perturb{\x}\in\reals^d$ is small enough for $\adv{\x}$ to be undistinguishable from $\x$ by a human, but causes the network to predict an incorrect label. 
 
Formally, given a trained network $g_\theta$ and a $p$-norm, the adversarial example is generated by solving the following optimization problem:
\begin{equation}
\label{eq:adv}
\tilde{\x} = \argmax_{\tilde{\x}:\normp{\tilde{\x}-\x}\leq \adveps}
\loss\big(\network_\theta(\tilde{\x}, \X^k), \by\big),
\end{equation}
where $\adveps$ represents the strength of the adversary, and $p$ is the norm value. In words, we would like to maximize, rather than minimize, the loss function between the prediction of $g_{\theta}$ on the adversarial example and the correct label under the constraint that the adversarial example is similar to the original example in $p$-norm.

Assuming the loss function $\loss$ is differentiable, the authors of \cite{shaham2015understanding} proposed to take the first order Taylor expansion of $\x\mapsto\loss(\network_\theta(\x, \X^k),
\by)$ to compute $\perturb{\x}$ by solving the following problem:
\begin{equation}
\label{eq:fgsm}
\tilde{\x} = \argmax_{\tilde{\x}:\normp{\tilde{\x}-\x}\leq \adveps}
\big(\nabla_{\x} \loss(\network_\theta(\x, \X^k),
\by)\big)^T (\tilde{\x}-\x)\,
\end{equation}
When $p=\infty$ the solution to the optimization problem is 
$$
\tilde{\x} = \x + \adveps \, \text{sign}(\nabla_{\x} \loss(\network_\theta(\x, \X^k), \by)),
$$ 
which corresponds to the \emph{fast gradient sign method} proposed in~\cite{goodfellow2014explaining}. In other words, generating adversarial examples following the fast sign gradient method involves with taking the sign of the gradients of the loss function with respect to the input, multiply it by a small fraction so it will be indistinguishable to a human and add it to the original example. 

Note that a single training example is composed of two parts; an enrollment set and a test utterance. To mimic a realistic model attack, we add the adversarial noise only to the test utterance and leave the enrollment set unchanged.

\section{Experiments}
\label{sec:experiments}

The setting was similar in all our experiments. We represented the speech utterances by acoustic features and trained a speaker verification model on that representation. Then we generated adversarial examples by adding noise to the feature vectors, and finally, acoustic waveforms were \emph{reconstructed}  from the adversarial example. For a reference, we also reconstruct waveforms of the original examples. The waveforms corresponding to the adversarial examples were used to fool the trained model as well as a different model that was trained under different conditions. We now describe the experimental setting in detail.

We evaluated the effectiveness of our method on two datasets: YOHO \cite{campbell1995testing} and NTIMIT \cite{jankowski1990ntimit}, each sampled at 8kHz. We used two sets of acoustic features. The first set of features was the Mel-spectrum. The acoustic signal was split into frames of 64 milliseconds with a shift of 4 milliseconds\footnote{The high shift frequency allows us a better reconstruction of the signal.}. Then we applied Hamming window and computed STFT. We used a 65 Mel frequency channels that yielded $d=65$ acoustic features. The second set we used was the Mel-Frequency Cepstrum Coefficients (MFCCs), extracted from the Mel-spectrum. Overall we trained four models; for each of the datasets (NTIMIT and YOHO), we used both feature sets. 

For the YOHO corpus, each training example was generated by picking one of the verification utterances and associating it with a set of 10 random enrollment utterances. Ten speakers were excluded from the training set and used as a test set. For the NTIMIT corpus, since there are only ten utterances per speaker, we generate each of the training examples by picking one utterance to be the verification example and four other utterances to be an enrollment set. In both datasets, we swapped the enrollment set in half of the examples to generate negative instances.

We generated adversarial examples using the fast gradient sign method by adding adversarial perturbation to the test utterance vector $\x$  using several epsilon values. We found that $\epsilon \in (0.2,0.3)$ caused the classifier to misclassify the adversarial examples with a high probability while keeping them remarkably similar to the original ones, a description of the adversarial examples evaluation process can be found in the next subsection. The models` performance was evaluated using the precision of correct classifications, and not using the standard equal error-rate (EER) since we aimed to show the effectiveness of adversarial attacks rather than comparing our model to other speaker verification systems.

\subsection{ABX Testing}
To validate that the generated adversarial examples are indeed indistinguishable by humans we performed an ABX test.  An ABX test is a standard way to assess the detectable differences between two choices of sensory stimuli. We presented to listeners two audio samples A and B; each being either the original (reconstructed) waveform or an adversarial waveform of the same example. These two samples are followed by a third sound X which was randomly chosen to be either A or B. The listener was instructed to decide whether X is more similar to sample A or sample B. We randomly sampled 50 pairs of audio examples, original and adversarial ones. All waveforms were reconstructed from Mel-spectrogram using the Griffin-Lim algorithm \cite{griffin1984signal}. eight different listeners tested each audio pair. Overall, on average 54\% of the examples were correctly classified by the human listeners. Subsequently, we use such indistinguishable adversarial examples to test the robustness of speaker verification system.

\subsection{A white-box attack}

In the setting of a white-box attack, we assume that the adversary has access to the internals of the model to be attacked. In other words, the attacker has complete knowledge and control of the network and can access the networks` gradients. An adversary can use these gradients to perturb the original input to become adversarial. The adversarial examples were crafted directly on the inputs that fed to the network. 

\begin{table}[t]\centering
\small
\renewcommand{\arraystretch}{0.8}
\def\sym#1{\ifmmode^{#1}\else\(^{#1}\)\fi}
\caption[Short Heading]{\protect System accuracy under white-box attacks.}\label{white_box}
\begin{tabular}{l*{4}{c}}
\toprule
\multicolumn{4}{c}{YOHO}\\
{}           &  Original test        &  Adversarial test        & Diff \\
\midrule
Mel-spectrum              &  85.50\%     &  37.50\%                 & 48.00\% \\
\midrule
MFCC             &  87.50\%     &  25.75\%                 & 61.75\% \\
\midrule
\multicolumn{4}{c}{NTIMIT}\\
{}          &  Original test        &  Adversarial test        & Diff \\
\midrule
Mel-spectrum              &  84.26\%     &  24.40\%                 & 59.86\% \\
\midrule
MFCC             &  82.14\%     &  10.20\%                 & 69.94\% \\
\bottomrule
\end{tabular}
\end{table}

Table~\ref{white_box} summarizes the results. The upper panel and the lower panel describes the results for the YOHO and the NTIMIT corpus, respectively. For each dataset, the accuracy on the test set is given in the first column. Even though one could achieve better results, our focus is to demonstrate the effectiveness of adversarial attacks on this model and to propose alternative ways (in addition to the traditional ones) of evaluating speaker verification systems. We assume the performance difference observed here is due to limited training data (YOHO has around 15.6 hours of speech, while the "OK Google" dataset contains 333 hours off speech \cite{heigold2016end}). In the second column, we presented the accuracy of the adversarial examples (generated from the test set examples). The last column is the degradation in performance. 

\begin{table}[t]\centering
\small
\renewcommand{\arraystretch}{0.8}
\def\sym#1{\ifmmode^{#1}\else\(^{#1}\)\fi}
\caption[Short Heading]{\protect False-positive rate (FPR) under white-box attacks.}\label{fpr}
\begin{tabular}{l*{3}{c}}
\toprule
\multicolumn{3}{c}{YOHO}\\
{}           &  Original test        &  Adversarial test \\
\midrule
Mel-Spectrum              &  1.46\%     &  69.76\%     \\
\midrule
MFCC             &  4.88\%     &  94.63\%             \\
\midrule
\multicolumn{3}{c}{NTIMIT}\\
{}          &  Original test        &  Adversarial test \\
\midrule
Mel-Spectrum              &  10.98\%     &  79.19\%     \\
\midrule
MFCC             &  1.73\%     &  82.08\%              \\
\bottomrule
\end{tabular}
\end{table}

In speech verification systems it might be considered more important to perform well regarding \emph{false-positive rate} (FPR), this is the case where an adversary claims to be someone else and is wrongfully accepted. Results of FPR are given in Table~\ref{fpr}, where we can see a significant degradation regarding FPR performance during the adversarial attack.

\subsection{Cross-dataset}

In the setting of black-box attack, the adversary has no access the model internals, only to its inputs and outputs. This setup is the strongest since it assumes no knowledge of the adversary regarding the type of model, its architecture or parameters. Moreover, the success of a black-box attack almost assures that a white-box will succeed equally or better. In our work,  we performed two back-box attacks: a \emph{cross-dataset} attack and a \emph{cross-feature} attack. In the case of cross-dataset attach, the adversarial examples were crafted using a model trained on dataset A, is used to attack a model trained on dataset B.

We trained two models: model A was trained on the YOHO dataset using MFCC features, while model B was trained on the NTIMIT dataset using Mel-spectrum features. Then, model B was used to create adversarial examples on NTIMIT; these examples were used to attack model A. In other words, models A and B were trained on two different datasets. The examples created by model B were used to attack model A. 

We found that model A reached an accuracy of 81.55\% on NTIMIT reconstructed clean test set and 58.93\% on NTIMIT reconstructed adversarial test set, a difference of 22.62\%. The FPR was degraded from 12\% to 46\%. 

\subsection{Cross-features}

A different type of back-box attack is done by creating adversarial examples using one set of acoustic features and attacking a model that was trained on a different set of acoustic features. More specifically, we trained a model on the YOHO corpus where the features were Mel-spectrum. We created adversarial examples and reconstructed waveforms. We then attacked a model that was trained on YOHO, but the features were MFCC. 

We found that the MFCC model reached an accuracy of 81\% on the reconstructed clean test set and accuracy of 62.25\% on the reconstructed adversarial test set with a difference of 18.75\%. The FPR degraded from 16\% to 46\%.

%
%
%
%

\section{Conclusion}
\label{sec:conclusion}

While deep neural networks have shown to improve the accuracy compared to the traditional speech verification components~\cite{snyder2017deep}, it becomes critical to revisit the evaluation protocol of those models and design new ways to assess their reliability beyond the traditional metrics. 

For future work, we would like to evaluate the robustness of the traditional speaker verification systems to adversarial examples as well as to apply adversarial training techniques to make the neural-based ones more robust to these type of attacks.


\bibliographystyle{IEEEbib}
\bibliography{refs}

\end{document}